\documentclass[english]{lni}

\usepackage{booktabs}
\usepackage{multirow}
\usepackage{graphicx}
\usepackage[]{blindtext}

\begin{document}
\title[Ein Kurztitel]{GenIC: An LLM-Based Framework for Instance Completion in Knowledge Graphs}
 \author[1]{Amel Gader}{amel.gader@uni-passau.de}{}
 \author[1]{Alsayed Algergawy}{alsayed.algergawy@uni-passau.de}{}

 \affil[1]{Chair of Data and Knowledge Engineering\\University of Passau\\Germany}
 
\maketitle

\begin{abstract}
Knowledge graph completion aims to address the gaps of knowledge bases by adding new triples that represent facts. The complexity of this task depends on how many parts of a triple are already known. Instance completion involves predicting the relation-tail pair when only the head is given (h, ?, ?).
Notably, modern knowledge bases often contain entity descriptions and types, which can provide valuable context for inferring missing facts. By leveraging these textual descriptions and the ability of large language models to extract facts from them and recognize patterns within the knowledge graph schema, we propose an LLM-powered, end-to-end instance completion approach.
Specifically, we introduce GenIC: a two-step \textbf{Gen}erative \textbf{I}nstance \textbf{C}ompletion framework. The first step focuses on property prediction, treated as a multi-label classification task. The second step is link prediction, framed as a generative sequence-to-sequence task. Experimental results on three datasets show that our method outperforms existing baselines.  Our code is available
at \href{https://github.com/amal-gader/genic}{https://github.com/amal-gader/genic}.

\end{abstract}
\section{Introduction}

Knowledge graphs are integral to various downstream applications, including question answering \cite{chakraborty2021introduction} \cite{saxena2020improving}, recommender systems \cite{guo2020survey}, and more recently, enhancing the predictions and interpretability of large language models \cite{pan2024unifying}. Given their critical role in these applications, improving the quality of knowledge graphs is crucial. A key metric for evaluating this quality is completeness. Unfortunately, all knowledge graphs suffer from varying degrees of incompleteness. For instance, in \href{https://www.example.com}{Wikidata}, one of the largest knowledge bases, only 0.5\% of football players have a high completeness rate (>96\% average completeness) \cite{balaraman2018recoin}. This issue has led to a plethora of research initiatives focused on the task of knowledge graph completion.
\\
\\
Knowledge graph completion comprises four main related tasks: Link prediction  which involves predicting either the tail entity or the head entity when given the relation and one other component (i.e predicting the tail given the head and relation (h, r, ?) or predicting the head given the relation and tail (?, r, t)). Relation prediction focuses on identifying the relation between a given head and tail entity (h, ?, t). Whereas instance completion takes on a greater challenge by predicting both the relation and the tail entity given only the head (h, ?, ?). More recently a new task entitled Triple set prediction \cite{zhang2024start} has emerged, aiming to achieve the ultimate goal of fully automating knowledge graph completion by predicting entire triples (?, ?, ?) without any predefined components or predicates.
\\
In this paper, we focus on the task of instance completion, which treats a fact or triplet as an instance. In this context, the given head entity serves as an incomplete sentence that we aim to complete. We argue that the instance completion task is a realistic and practical approach, as it requires only one known component of the triplet. This makes it particularly useful for entities with few connections or sparse associated facts.

Our primary goal is to leverage the rich textual information embedded within knowledge graphs, such as descriptions and entity types, by transforming the instance completion task into a natural language processing problem. Specifically, we take the first step to solve this task using, as the foundation, pre-trained language models. Our method adopts a two-step pipeline: first, property prediction is treated as a multi-label classification problem, and second, link prediction is modeled as a sequence-to-sequence task, following the approach of \cite{xie2022discrimination}.

\section{Related Work}
Knowledge Graph completion seeks to enhance knowledge bases by adding missing facts. Numerous studies have addressed this task through various approaches, including link prediction, relation prediction, and instance completion. A growing body of research leverages pretrained large language models to tackle KG completion. In the following, we provide a brief overview of related work, focusing specifically on instance completion, as well as recent work that employs LLMs to enrich KGs.
\subsection{Instance Completion}
The task of instance completion involves associating a relation-tail pair with a given head entity. One notable approach for this task is RETA \cite{rosso2021reta}, which predicts relation-tail pairs for a head entity in two steps. First, a filter identifies potential relation-tail candidate pairs by leveraging schema information from knowledge graphs, generating possible triplet-type combinations and their frequencies. In the second step, these predicted triplets are ranked using schema patterns and structural information through a language embedding model.
OKELE \cite{cao2020open} is another instance completion solution, comprising three steps. It begins with an attention-based GNN model for property prediction, followed by link prediction using external web sources. Lastly, a probabilistic approach is employed for fact verification. Unlike our method, which focuses on internal entity linking within the knowledge base, OKELE enriches knowledge by incorporating external web data.
\subsection{LLMs for Knowledge Graphs}
KG-BERT \cite{yao2019kg} was the first to approach triple plausibility as a sequence classification task using pretrained language models. It has been applied to tasks such as triple classification, relation prediction, and link prediction, achieving state-of-the-art results. However, KG-BERT requires negative sampling and ranks all entities to identify the best candidates, which is computationally expensive and resource-intensive. To address these limitations, GenKG \cite{xie2022discrimination} generates only the top-k candidate entities through entity-aware hierarchical decoding, significantly reducing computational overhead.
KEPLER \cite{wang2021kepler} proposes a unified model that combines the rich textual information from knowledge graphs with their structural and relational facts. It achieves this by jointly training a knowledge embedding model and a masked language model, optimizing both objectives simultaneously.
KG-LLM \cite{shu2024knowledge} leverages large language models to enhance multi-hop link prediction by transforming knowledge graph paths into natural language prompts. These prompts describe the sequence of relations between interconnected nodes in the graph. By following the provided instructions, the model infers the interrelations and identifies missing links, effectively predicting paths between entities based on the structure of the knowledge graph.
For further exploration of large language model (LLM)-enhanced knowledge graphs, we refer the reader to this relevant survey \cite{pan2024unifying}.

\section{Method}
\subsection{Overview of the proposed approach}
A knowledge graph \( \mathcal{G} = (\mathcal{E}, \mathcal{R}, \mathcal{F}, \mathcal{C}, \mathcal{D}) \)
 is a knowledge structured representation consisting of entities, relations, and facts. Each fact \( f \in \mathcal{F} \)
 is a triplet in the form \( (h,r,t) \in \mathcal{E} \times \mathcal{R} \times \mathcal{E} \), where \(h\) (head) and  \(t\) (tail) are elements from the set of entities $\mathcal{E}$  and \(r\) (relation) is an element from the set of relations $\mathcal{R}$. Entities can be associated with a type or class \( c \in \mathcal{C} \)
 and a textual description \( d \in \mathcal{D} \), where $\mathcal{C}$ and $\mathcal{D}$ represent the sets of categories and descriptions corresponding to the entities within the knowledge graph. 
 Traditional knowledge graph embedding models do not make advantage of $\mathcal{C}$ and $\mathcal{D}$, our goal in this paper is to leverage this information through pretrained language models to solve the instance completion task.

A straightforward approach to solving the instance completion problem is to evaluate all possible relation-tail combinations for a given head. The number of candidates for each head theoretically scales with $|\mathcal{R}| \times |\mathcal{E}|$, where $|\mathcal{R}|$ represents the number of relations and $|\mathcal{E}|$ the number of entities in the knowledge graph. For example, with just 1,000 entities from the FB15k-237 dataset \cite{toutanova-chen-2015-observed}, the number of potential candidates per head would be 
$237 \times 14.5 \times 10^3$, resulting in $\sim 3 \times 10^6$ candidate pairs and leading to $\sim 3 \times 10^9$ possible triplets, a daunting task to rank effectively.

\cite{rosso2021reta} Addresses this by reducing the candidate search space to $|\mathcal{T}_h| \times |\mathcal{R}| \times |\mathcal{T}_t|$, where $\mathcal{T}_t$ and $\mathcal{T}_t$ denote the sets of head and tail types, respectively, and $\mathcal{R}$ represents the set of relations in the graph. In practice, $\mathcal{T}_t$ and $\mathcal{T}_h$ are equivalent, representing all entity types. However, this method is unable to leverage entity descriptions, and achieves a modest performance. As an alternative, we suggest to divide the task into two steps, which drastically reduces its complexity.

\begin{figure}
  \centering
  \includegraphics[width=.8\textwidth]{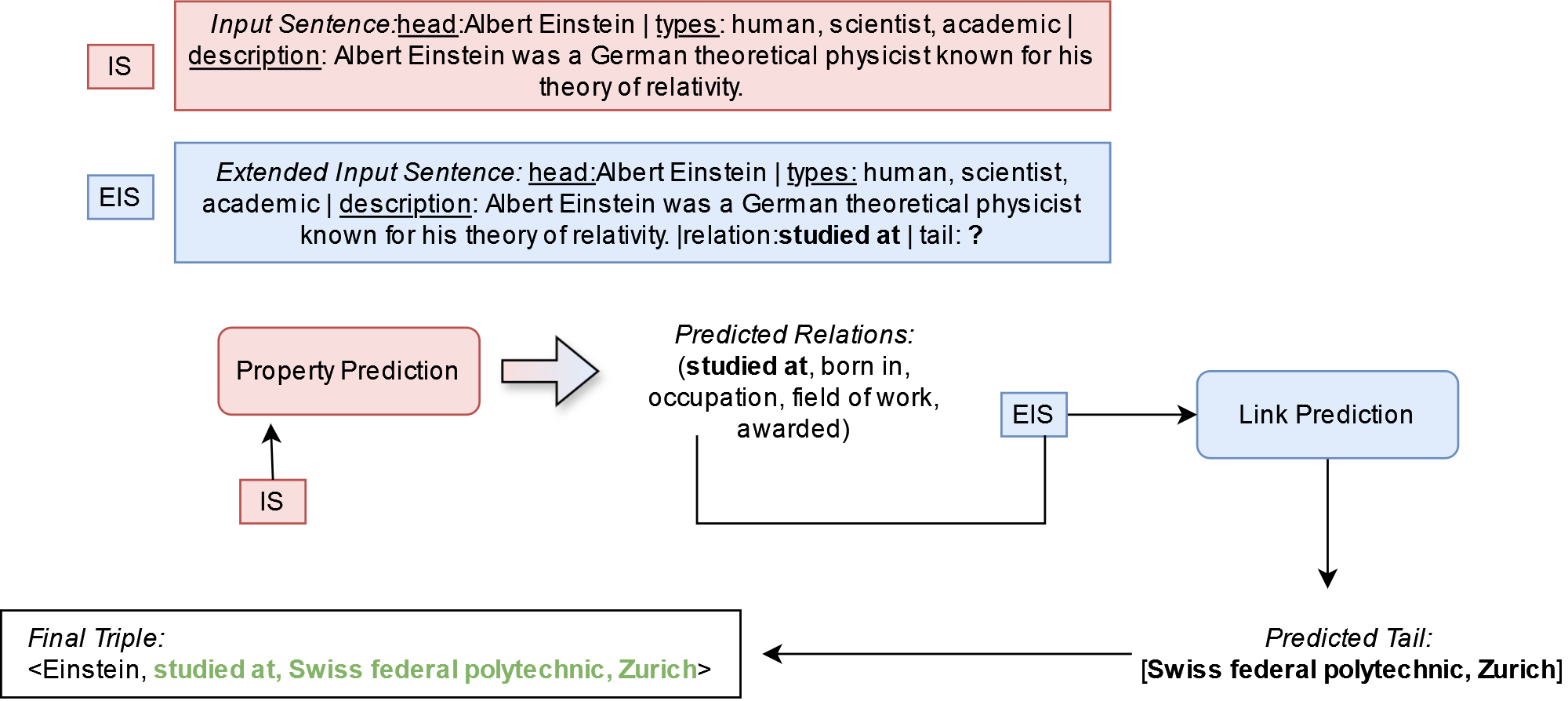}
  \caption{Figure depicting our Full-fledged pipeline from property prediction to link prediction.}
  \label{fig:pipeline}
\end{figure}

The first step is relation prediction, where for each head entity, we predict a set of relevant properties. Based on this predicted set, we then create (head, relation) pairs, which are used as input for the subsequent link prediction model as presented in figure \ref{fig:pipeline}. 

To prevent data leakage, we ensure that the same data splits are used consistently in both steps. This guarantees that any (h, r) pair generated by the property prediction step during the test phase is entirely new to the link prediction model.

The input for both steps is a text sequence, where entities, relations, and descriptions are combined and treated as continuous text. The first step in processing these sequences involves tokenization, which converts the text into a sequence of tokens \((t_1,…,t_n)\) These tokens are then encoded by projecting them into the model’s semantic space. The resulting representations are used to predict a set of properties in the first step, and a sequence of tokens (tail entity) in the second step.

Entity types are incorporated into the model to filter out irrelevant properties. For instance, if the entity type is \texttt{human}, the model can focus on properties typically associated with humans. For more specific types, such as \texttt{artist}, the model can refine its predictions to include relationships relevant to that class. Additionally, entity descriptions help predict missing or new links by providing contextual information.

Furthermore, Large language models are capable of capturing semantic similarities and leveraging their pre-trained general knowledge, which is particularly valuable for guiding predictions. For example, the name of an entity can provide clues about its type, such as a person's name indicating that the entity is human. This built-in knowledge enables the model to make more informed predictions even in cases where explicit type information is unavailable.

\subsection{Property Prediction}
The task of property prediction involves identifying a set of relevant properties or relations for a given head entity h. 
This can be expressed as predicting the set $\{r_1, r_2, r_3, \dots, r_k\}_h$.

We model this task as a multi-label classification problem. For each given head entity, we predict a binary vector \((0, 0, 1, \dots, 1)\), where a \(1\) at position \(k\) indicates that relation \(r_k\) is relevant to the head, and a \(0\) denotes irrelevance.
\\
We leverage a pretrained language model, specifically the Mistral architecture \cite{jiang2023mistral}, for this task. The input to our model is a sequence comprising the head entity, its type, and its description, all represented as a sequence of tokens. The model outputs a binary vector, where each element corresponds to a relation from the knowledge graph’s set of relations, with the output vector's dimension equal to $|\mathcal{R}|$, the total number of relations in the graph.
To optimize the model, we use the binary cross-entropy loss function, defined as:

\[
L = -\frac{1}{N} \sum_{i=1}^{N} \sum_{c=1}^{C} \left[ y_{i,c} \log(\hat{y}_{i,c}) + (1 - y_{i,c}) \log(1 - \hat{y}_{i,c}) \right]
\]

where:
\begin{itemize}
    \item \(N\) is the total number of samples, \(C\) is the number of labels (relations).
    \item \(y_{i,c}\), \(\hat{y}_{i,c}\) are respectively the true label and the predicted probability for the \(i\)-th sample and \(c\)-th class.
\end{itemize}

Incorporating type information is crucial for the property prediction task, as entities of the same class are likely to share common relations. For instance, all humans may share relations like \texttt{birthplace}, \texttt{birthdate}, \texttt{nationality}, \texttt{education}, and \texttt{occupation}. Furthermore, within the broader class of humans, more specific subclasses based on occupation for example, can refine the precision of predicted relations. An artist or a scientist, for instance, might have additional distinctive properties such as \texttt{awards}, \texttt{field of work}, \texttt{publications}, or \texttt{artwork}. That’s why we follow \cite{balaraman2018recoin} and use the occupation if available as an entity type, to add more context. The description is equally crucial, as it can contain complete facts or explicit hints about key relationships. In cases where an entity's description and type are missing, the model may base its predictions on semantic similarities and the head mention to infer the relevant properties.


\subsection{Link Prediction}
We model the link prediction task as a sequence-to-sequence problem, leveraging an encoder-decoder architecture, specifically using the T5 model \cite{raffel2020exploring}. The input sequence for link prediction is similar to that used in property prediction, with the addition of the relation. The output is a sequence of tokens that represent the tail entity.
Given a head entity \(h\), its class or type \(c\), a description \(d\), and a relation \(r\), we structure the input prompt as:\textit{"head: h, types: c, description: d, relation: r, tail: …."}, as depicted in figure \ref{fig:pipeline}. The model then autoregressively generates t, the predicted tail entity.

The contextual information provided by the description and entity type can offer either explicit or implicit hints. In cases where the fact is directly mentioned in the description, the task resembles information extraction from textual data. In other cases, the context serves as an informative guide for predicting the tail entity's category. The combination of the head type and relation forms a useful pattern that the model can leverage to improve its predictions. For example, if the head entity type is \texttt{academic} and the relation is \texttt{studied at}, the predicted entity is likely to be of type \texttt{university} or \texttt{higher institute}, or among similar types.

We optimize using the standard cross-entropy loss: 
\[
L = -\frac{1}{N} \sum_{i=1}^{N}  \log(P(y_{i-1}, \ldots, y_2, y_1))
\]
where \(P(y_{i-1}, \ldots, y_2, y_1)\) is the predicted conditional probability of token \(y_i\) given the previous predicted tokens.

As in \cite{xie2022discrimination} we use beam search for decoding to generate the top-k predictions, offering a more efficient alternative to traditional link prediction methods like TransE \cite{bordes2013translating} or RotatE \cite{sun2019rotate}, which require ranking all possible entities.

\section{Experiments}
\textit{Experimental Setup.} Our experiments were conducted on an Intel Xeon Gold 6338 CPUs with 500 GB of memory, paired with NVIDIA A100 80G GPUs.
In our experiments we try to compare our \textbf{GenIC} performance against the baselines and we conduct an ablation study to understand the key features of the model.

\textit{Datasets.} We evaluated our approach using three RDF knowledge bases as benchmarks.
FB15k-237, extracted from Freebase, is a refinement of the FB15k \cite{bordes2013translating} by removing inverse relations that could lead to data leakage, as identified by \cite{toutanova-chen-2015-observed}.
CoDEx-l is a knowledge graph completion dataset derived from \href{https://www.example.com}{Wikidata} and Wikipedia, introduced in \cite{safavi-koutra-2020-codex}.
A subset of \href{https://wordnet.princeton.edu/}{Wordnet}, WN18RR \cite{dettmers2018convolutional}, is an improved version of the original WN18 \cite{bordes2013translating} dataset, specifically curated to exclude inverting triples that could constitute data leakage. Details about the datasets are outlined in table \ref{tab:dataset_statistics}.

\begin{table}[h!]
\centering
\begin{tabular}{|c|c|c|c|c|c|c|}
\hline
\textbf{Dataset} & \textbf{\#Ent} & \textbf{\#Rel} & \textbf{\#Facts} & \textbf{\#Train} & \textbf{\#Val} & \textbf{\#Test} \\ \hline
FB15k-237 & 14,541 & 237 & 310,116 & 217,081 & 46,517 & 46,517 \\ \hline
WN18RR & 40,943 & 11 & 93,003 & 65,102 & 13,950 & 13,950 \\ \hline
CoDeX & 77,951 & 69 & 612,437 & 428,705 & 91,865 & 91,865 \\ \hline
\end{tabular}
\caption{Datasets Statistics}
\label{tab:dataset_statistics}
\end{table}

\textit{Pre-processing.} For the multi-label classification task, addressing class imbalance is crucial. To ensure balanced representation across data splits, we first load all data, shuffle it, and then perform a stratified split into training, validation, and test sets. The stratification is based on the head type, which is a key factor in determining properties. This helps in maintaining consistent class distribution across splits.

\textit{Baselines.} For the property prediction step we implemented two baseline techniques. Recoin is a method to evaluate the completeness of entities in \href{https://www.example.com}{Wikidata} by comparing them to similar entities and identifying relevant missing properties based on their frequency within the entity's class. It calculates a weighted frequency score for properties, taking into account the entity's membership in multiple classes. For example, if an entity belongs to both the \texttt{capital} and \texttt{city} classes (e.g Berlin), the Recoin score for a property such as \texttt{population} is given by: \[
\text{Score}_{\text{population}} = \frac{freq_\text{population in city} + freq_\text{population in capital}}{\text{Size}_{\text{city}} + \text{Size}_{\text{capital}}}
\]
A Hybrid Recommendation Model; we employ a hybrid approach that combines item-based K-nearest neighbors (item-KNN) \cite{sarwar2001item} with content-based filtering. The item-KNN model, a collaborative filtering technique, predicts the properties of an entity by averaging the properties of its closest neighbors based on existing attributes. Complementing this, the content-based filtering method leverages textual information, such as the entity's description and type using a TF-IDF vectorizer. This process generates a cosine similarity vector, comparing the entity with others in the dataset. Entities are represented as binary vectors of properties, allowing the model to integrate both collaborative and content-based predicted properties. For the link prediction step we used the implementation provided by \cite{sun2019rotate} of TransE and RotatE.
TransE \cite{bordes2013translating} is a canonical model that considers relations as translations $h + r \approx t$.
While RotatE \cite{sun2019rotate} considers the relation as a rotation from head to tail in the complex space, which makes it more efficient in inferring more complex and implicit relations. 

For a complete instance completion pipeline we combine a property prediction baseline method with one of the link prediction techniques as outlined in table \ref{tab:results_ic}.

\textit{Evaluation metrics.}
For the property prediction step, we use standard multi-label classification metrics, including micro F1 score, Recall, and Precision. For the link prediction task, we evaluate performance using the commonly adopted metrics: Hits@1, Hits@5, and Hits@10. These metrics indicate how often the correct answer appears within the top-k ranked predictions, where $k \in {1,5,10}$, respectively. Whereas for instance completion we use the same metrics of the link prediction as its the last step of the task coupled with the precision of predicted h-r candidate pairs from the first step.

\textit{Models and settings} : As mentioned earlier, we utilize the Mistral model for the property prediction step, specifically the Hugging Face implementation \href{https://huggingface.co/mistralai/Mistral-7B-v0.1}{\texttt{mistral-7B}}. We fine-tune the model using parameter-efficient techniques, retraining only 0.19\% (13.7 million out of 7.1 billion) of the original parameters, while employing quantization to minimize memory usage. For link prediction, we leverage the large checkpoint of the T5 model, also implemented via \href{https://huggingface.co/google-t5/t5-large}{Hugging Face}. We reduce the number of trainable parameters to 0.64\%, training only 4.7 million parameters, by applying LoRa parameter-efficient fine-tuning \cite{hu2021lora}.

\textit{Results.}
Table \ref{tab:results_pp} depicts the results of the property prediction step. Our method shows a notable improvement in F1-score, ranging from 7-11\% across all three datasets when compared to Recoin. This gain demonstrates the effectiveness of our approach in identifying relevant properties more accurately thanks to the rich textual information and its embedded knowledge. The recommendation model underperforms, likely because it relies on a single property to generate suggestions. We use only one connection, as the minimum requirement for an entity to be included in the knowledge graph is having at least one link.

In the case of the CoDEx dataset, we observe a substantial increase in precision, while the recall remains comparable to that of Recoin. This suggests that our model is better at reducing false positives, unlike Recoin’s heuristic-based method, which may predict irrelevant properties based on the class or type.

\begin{table}[ht]
\centering
\caption{Property Prediction results on CoDEx, FB15k-237 and WN18RR.}
\resizebox{\textwidth}{!}{%
\begin{tabular}{cccccccccc}
\toprule
\multirow{2}{*}{Method} & \multicolumn{3}{c}{\textbf{CoDEx}} & \multicolumn{3}{c}{\textbf{FB15K-237}} & \multicolumn{3}{c}{\textbf{WN18RR}} \\ 
\cmidrule(lr){2-4} \cmidrule(lr){5-7} \cmidrule(lr){8-10}
                        & Precision & Recall & F1 Score  & Precision & Recall & F1 Score  & Precision & Recall & F1 Score  \\ 
\midrule
Hybrid Recommender                & 0.594      & 0.471   & 0.490   & 0.595      & 0.492   & 0.497  & 0.667      & 0.482   & 0.520     \\ 
Recoin                & 0.708      & 0.778   & 0.724      & 0.718      & 0.781   & 0.791      & 0.790      & 0.713   & 0.716      \\ 
GenIC-PP              & \textbf{0.837}      & \textbf{0.809}   & \textbf{0.823}      & \textbf{0.868}      & \textbf{0.870}   & \textbf{0.869}      & \textbf{0.844}      & \textbf{0.814}   & \textbf{0.829}      \\ 
\bottomrule
\end{tabular}%
}

\label{tab:results_pp}
\end{table}

Table \ref{tab:results_lp} presents the results of the link prediction step. Our model demonstrates improvement over the baselines on the CoDEx dataset, achieving a 17.7\% increase in accuracy (Hits@1). This improvement can be attributed to the differing characteristics of the datasets. CoDEx, being derived from Wikidata, aligns well with the language model’s familiarity with the underlying textual structure and encyclopedic knowledge. In contrast, WN18RR, based on WordNet, emphasizes relational and hierarchical interconnections, where textual information plays a less significant role compared to structured relationships. Additionally, in the case of CoDEx, we use \texttt{occupation} as a type, which provides the model with a stronger contextual clue, increasing the likelihood of ranking the correct link within the top ten with a 0.593 probability. 

\begin{table}[ht]
\centering
\caption{Link Prediction results on CoDEx, FB15k-237 and WN18RR.}
\resizebox{\textwidth}{!}{%
\begin{tabular}{cccccccccc}
\toprule
\multirow{2}{*}{Method} & \multicolumn{3}{c}{\textbf{CoDEx}}& \multicolumn{3}{c}{\textbf{FB15K-237}} & \multicolumn{3}{c}{\textbf{WN18RR}} \\ 
\cmidrule(lr){2-4} \cmidrule(lr){5-7} \cmidrule(lr){8-10}
                        & Hits@1 & Hits@5 & Hits@10  & Hits@1 & Hits@5 & Hits@10  & Hits@1 & Hits@5 & Hits@10  \\ 
\midrule
TransE                & 0.208      & 0.376   & 0.432      & 0.230 & 0.436   & 0.528      & 0.0130      & 0.462   & 0.528      \\ 
RotatE                & 0.257      & 0.388   & 0.436      & \textbf{0.240}      & \textbf{0.439}   & \textbf{0.529}      & \textbf{0.427}      & \textbf{0.525}   & \textbf{0.571}      \\ 
GenIC-LP              & \textbf{0.434}      & \textbf{0.545}   & \textbf{0.593}      & 0.142      & 0.263   & 0.337      & 0.277      & 0.456  & 0.521      \\ 
\bottomrule
\end{tabular}%
}
\label{tab:results_lp}
\end{table}

The synthetic results in Table \ref{tab:results_ic} demonstrate the effectiveness of GenIC across all three datasets, with the exception of WN18RR. On WN18RR, the Hits@5 and Hits@10 scores using the Recoin-GenIC-LP combination are slightly higher than those of GenIC. However, it's important to note that this comparison isn't entirely valid, as the predicted h-r pairs differ between the two approaches in the first step. Additionally, we observe better precision when using the GenIC-PP model. Here, precision refers to the percentage of true positive h-r pairs among the predictions, indicating the performance of the first step. Overall, we can conclude that our method outperforms the baselines.

\begin{table}[ht]
\centering
\caption{Instance Completion results on CoDEx, FB15k-237 and WN18RR.}
\resizebox{\textwidth}{!}{%
\begin{tabular}{ccccccccccccc}
\toprule
\multirow{2}{*}{Method} & \multicolumn{4}{c}{\textbf{CoDEx}} & \multicolumn{4}{c}{\textbf{FB15K-237}} & \multicolumn{4}{c}{\textbf{WN18RR}} \\ 
\cmidrule(lr){2-5} \cmidrule(lr){6-9} \cmidrule(lr){10-13}
                        & Hits@1 & Hits@5 & Hits@10 &covergae & Hits@1 & Hits@5 & Hits@10 &coverage & Hits@1 & Hits@5 & Hits@10 &Precision \\ 
\midrule
Recoin + TransE                      & 0.0350   & 0.191  & 0.247 & 0.892   & 0.112      & 0.275   & 0.340  & 0.901      & 0.0657   & 0.180 & 0.208    & 0.885    \\ 
 Recoin + RotatE    & 0.0714      & 0.251   & 0.286  &   0.892          & \textbf{0.140}      & 0.282   & 0.347  & 0.901        & 0.159      & 0.198   & 0.211     & 0.885   \\ 
  Recoin + GenIC-LP                & 0.282      & 0.539   & 0.606 & 0.892     & 0.0876      & 0.183   & 0.229    & 0.901    & 0.253      & \textbf{0.429}   & \textbf{0.495}     & 0.885   \\ 
   GenIC-PP + TransE                & \textbf{0.360}      & 0.460   & 0.493   & \textbf{0.922} & 0.132      & 0.306   & 0.374   & 0.957     & 0.0396      & 0.307   & 0.350     & \textbf{0.896}   \\ 
GenIC              & 0.311      &\textbf{0.540}   & \textbf{0.613} &\textbf{0,922}     & 0.127      & \textbf{0.311}   & \textbf{0.381}   & \textbf{0.957}    & \textbf{0.267}      & 0.425   & 0.484  & \textbf{0.896}      \\ 
\bottomrule
\end{tabular}%
}
\label{tab:results_ic}
\end{table}

The ablation study in table \ref{tab:results_ablation} highlights the importance of both textual and schema information, particularly an entity's description and type. Even when one component is absent, the other often provides sufficient context to sustain strong performance; for example, the model that includes descriptions but no types still performs well across all three tasks. In contrast, the baseline model without either types or descriptions demonstrates a drastic decline in performance, showing a 27.5 percentage point decrease in link prediction accuracy and an 18.6 point loss in F1 score for the property prediction task. Additionally, there is a notable decrease in instance completion performance.
\begin{table}[ht]
\centering
\caption{Ablation study results on CoDEx.}
\resizebox{\textwidth}{!}{%
\begin{tabular}{ccccccccccc}
\toprule
\multirow{2}{*}{Task} & \multicolumn{3}{c}{\textbf{Link Prediction}} & \multicolumn{3}{c}{\textbf{Property Prediction}} & \multicolumn{3}{c}{\textbf{Instance Completion}} \\ 
\cmidrule(lr){2-4} \cmidrule(lr){5-7} \cmidrule(lr){8-11}
                        & Hits@1 & Hits@5 & Hits@10  & Precision & Recall & F1 Score  & Hits@1 & Hits@5 & Hits@10 & Precision \\ 
\midrule
w/o types                & 0.426      & 0.538   & 0.585      & 0.656      & 0.648   & 0.652      & 0.275      & 0.535   & 0.605 & 0.807    \\ 
w/o description                 & 0.421      & 0.537      & 0.579 & 0.773      & 0.703  & 0.736      & 0.246      & 0.442   & 0.530  & \textbf{0.947}   \\ 
w/o types, w/o description                      & 0.159   & 0.316      & 0.399 & 0.656     & 0.620   & 0.637      & 0.251      & 0.458   & 0.519 &  0.400   \\ 
with types \& description                      & \textbf{0.434}   & \textbf{0.545}      & \textbf{0.593} & \textbf{0.837}     & \textbf{0.809}   & \textbf{0.823}      & \textbf{0.311}      & \textbf{0.540}   & \textbf{0.613}  &  0.925  \\ 
\bottomrule
\end{tabular}%
}
\label{tab:results_ablation}
\end{table}
The performance variations observed in the property prediction task across all three ablated models suggest that textual and schema information are particularly vital in the property prediction step, as type and description serve as key features for identifying relevant shared properties and relationships across different categories.

\section{Limitations}

In this section, we outline the limitations of our work and suggest potential future directions. 
Firstly, our approach operates under the assumption that the datasets used in our experiments are complete, meaning entities are fully linked without any missing connections. This assumption is flawed, as it leads us to filter out all new links beyond the existing triplets, potentially misclassifying some true positives as false positives. This choice was made  because evaluating entirely new facts or links would necessitate external validation to verify their accuracy. Similarly, link prediction requires external validation to assess the correctness of newly predicted triples.

As a direction for future work, we propose incorporating a verification step utilizing large language models to validate the accuracy of predicted facts.

In addition, relying uniquely on language models may cause us to overlook the rich structural and relational information inherent in knowledge graphs. Therefore, one possible future direction is to explore a joint approach that integrates both language models and the structural features of knowledge graphs.
\section{Conclusion} 
In this paper, we propose an end-to-end LLM-powered solution for instance completion, advancing the automation of knowledge base completion to a more practical level.
By leveraging large language models and the rich textual information in knowledge bases like entity types and descriptions, our approach demonstrates the potential of language models in this realm. We achieve significant improvements over existing baselines on the separate property and link prediction steps as well as the overall instance completion pipeline, offering a more effective and realistic solution that holds great potential for future developments in knowledge graph enrichment.


\bibliographystyle{plain}
\bibliography{lni-paper-example-de}

\end{document}